\documentclass[10pt,twocolumn,letterpaper]{article}

\usepackage{iccv}
\usepackage{times}
\usepackage{epsfig}
\usepackage{graphicx}
\usepackage{amsmath}
\usepackage{amssymb}
\usepackage{import}
\usepackage{caption}
\usepackage{comment}
\usepackage{booktabs}
\usepackage{soul}
\usepackage{footnote}
\usepackage{comment}
\usepackage[accsupp]{axessibility}

\usepackage[breaklinks=true,bookmarks=false]{hyperref}

\iccvfinalcopy 

\def\iccvPaperID{10105} 

\begin{document}


\title{Cloth2Body: Generating 3D Human Body Mesh from 2D Clothing}
\author{Lu Dai$^{1,2^\ast}$, Liqian Ma$^{2^\dagger}$, Shenhan Qian$^{3}$, Hao Liu$^{1}$, Ziwei Liu$^{4^\dagger}$, Hui Xiong$^{1^\dagger}$\\
$^{1}$The Hong Kong University of Science and Technology (Guangzhou)\\
$^{2}$ZMO AI Inc.\\
$^{3}$Technical University of Munich\\
$^{4}$S-Lab, Nanyang Technological University\\
{\tt\small ldaiae@connect.ust.hk}
{\tt\small liqianma.scholar.outlook.com}
{\tt\small shenhan.qian@tum.de} \\
{\tt\small ziwei.liu@ntu.edu.sg}
{\tt\small \{liuh,xionghui\}@ust.hk}
}

\twocolumn[{
\renewcommand\twocolumn[1][]{#1}%
\maketitle
\begin{center}
    \centering
    \vspace{-20pt}
    \captionsetup{type=figure}
    \includegraphics[width=0.98\textwidth]{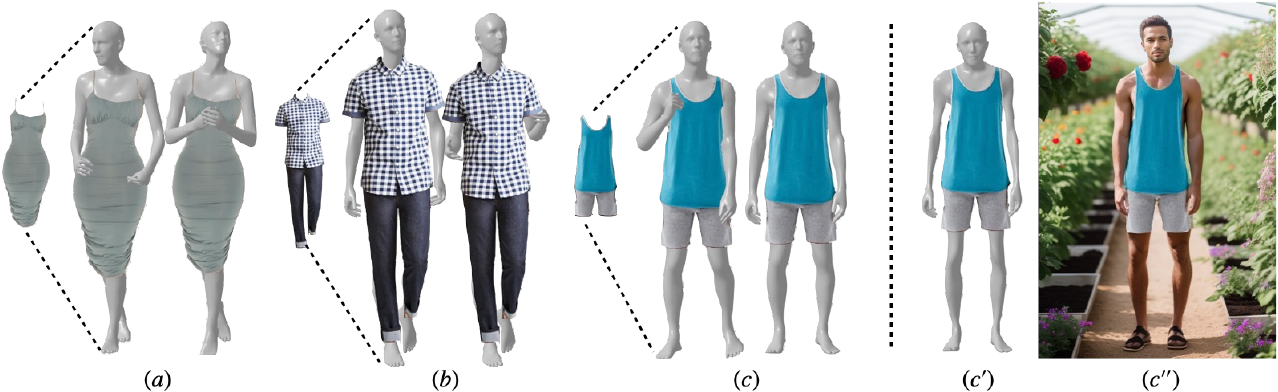}
    \captionof{figure}{\fontsize{8.5pt}{8.5pt}\selectfont In our Cloth2Body problem, our model takes a 2D clothing image as input and produces 3D human body of various poses, which can fit into the cloth pixel-wise when rendered back to the image plane, as shown in (a),(b) and (c). This framework also allows users to manipulate body figure within cloth size constraints. ($\text{c}^\prime$) is a shape variation from (c) but still in the same clothing and pose. ($\text{c}^{\prime\prime}$) is an application of our methods where a stable diffusion module consumes our output and generates human image.}
\end{center}%
}]


\renewcommand{\thefootnote}{\fnsymbol{footnote}}
\footnotetext[1]{Work partially conducted during an internship at ZMO AI Inc.}
\footnotetext[2]{Corresponding author}

\begin{abstract}
In this paper, we define and study a new \textbf{Cloth2Body} problem which has a goal of generating 3d human body meshes from a 2D clothing image. Unlike the existing human mesh recovery problem, Cloth2Body needs to address new and emerging challenges raised by the partial observation of the input and the high diversity of the output. Indeed, there are three specific challenges. First, how to locate and pose human bodies into the clothes. Second, how to effectively estimate body shapes out of various clothing types. Finally, how to generate diverse and plausible results from a 2D clothing image. To this end, we propose an end-to-end framework that can accurately estimate 3D body mesh parameterized by pose and shape from a 2D clothing image. Along this line,  we first utilize \textbf{Kinematics-aware Pose Estimation} to estimate body pose parameters. 3D skeleton is employed as a proxy followed by an inverse kinematics module to boost the estimation accuracy. We additionally design an adaptive depth trick to align the re-projected 3D mesh better with 2D clothing image by disentangling the effects of object size and camera extrinsic. Next, we propose \textbf{Physics-informed Shape Estimation} to estimate body shape parameters. 3D shape parameters are predicted based on partial body measurements estimated from RGB image, which not only improves pixel-wise human-cloth alignment, but also enables flexible user editing. Finally, we design \textbf{Evolution-based pose generation method}, a skeleton transplanting method inspired by genetic algorithms to generate diverse reasonable poses during inference.
As shown by experimental results on both synthetic and real-world data, the proposed framework achieves state-of-the-art performance and can effectively recover natural and diverse 3D body meshes from 2D images that align well with clothing.
\end{abstract}
\vspace{-10pt}
\section{Introduction}
\label{sec:intro}

3D virtual human is widely used in contemporary 
 industry, such as animation and game~\cite{avatarclip:22, dreamfusion:22}, VR/AR applications~\cite{vr:06} and fashion design~\cite{garment4d:21, garmentcollision:21,vto:19}. 3D human representation has advantage over 2D images due to its manipulation flexibility and generation robustness~\cite{eg3d:22}. Indeed, many studies have focused on recovering 3D humans from 2D images ~\cite{meshtransformer:21,hmr:18,hmrreview:22}. However, there are some application scenarios where target human is not present in an image but we still want to imagine the person from certain context, such as generating human face from hair or dressing~\cite{comodgan:21, pgpig:17} and generating  posed human in a scene~\cite{wang2021scene,yi2022human,nie2021pose2room}.

In this paper, we are interested in the scenario of generating human body meshes from a 2D clothing image. 
Existing methods can generate 2D human via deep generative models~\cite{mae:21,gfla:2020,pgpig:17}. However, these methods lack robustness and explainability. The generated 2D humans are not amenable to interactive manipulation afterwards, thus not satisfying the need of many industrial scenarios.
Therefore, we formulate a novel task called \textit{Cloth2Body} which aims to generate 3D virtual human mesh that can fit into 2D clothing image. This task is non-trivial as it faces three challenges:
\textbf{1) Partial observation:} Human body pixels are absent from the 2D clothing image and the mesh should be inferred from its interaction with clothing. 
\textbf{2) Pixel-wise alignment:} 3D human body should be well-aligned with 2D clothing in the pixel-wise level when re-projected onto the 2D plane. 
\textbf{3) Diverse outputs:} The same 2D clothing is potentially suitable for multiple 3D human bodies of different poses and shapes, so we need to model this outcome diversity.

To address these challenges, we propose an effective end-to-end framework.  
First, {\bf partial observation} compels the model to effectively exploit contexts and priors for making accurate predictions. As a solution, our pipeline utilizes spatial information and pose database as priors for invisible joints and exploits RGB context to accurately locate visible joints.
In addition, we explicitly estimate body measurements for shape estimation which significantly improves the model performance. Here, body measurements can either be extracted from the image by cloth landmark detection~\cite{dflmk:16} for convenience or from user input for interactive manipulation. The explicitly estimated measurements improve the model both in estimation accuracy and explainability. Second, {\bf pixel-wise alignment} requires the model to accurately localize 3D body in camera space. We adopt an inverse projection method and introduce a novel adaptive-depth projection trick to improve the alignment between the reconstructed joints and the inversely projected ones. Unlike camera parameter regression methods~\cite{hmr:18, expose:20}, our mechanism can guarantee image space alignment between the 2D clothing and the projected 3D body given accurate pose and shape. Finally, {\bf diverse output} aims to diversify out-of-cloth poses while preserving human-cloth alignment. We propose an evolution-based pose generation method that uses skeleton crossover and mutation to generate diverse reasonable 3D bodies. This training-free method is simple but effective for our pose diversifying purpose. Our method demonstrates superior results over other alternative methods adapted to this task.

To summarize our contributions: \textbf{1)} We propose a novel task Cloth2Body aiming at generating 3D human body meshes from a 2D clothing image. Many fashion-related downstream applications can benefit from 
 this task. \textbf{2)} We design an end-to-end framework that can effectively handle the new challenges emerging from the Cloth2Body problem setting. Specifically, our Landmark2Shape method and the adaptive-depth projection approach can also benefit other tasks such as human mesh recovery (HMR) towards better 2D alignment and explainability. \textbf{3)} We introduce two new datasets based on existing 3D human datasets for Cloth2Body training and evaluation.
\section{Related Work}
\label{sec:rw}
\noindent\textbf{Virtual Try-On.}
Virtual try-on aims at re-targeting clothes to a given person. 2D-based virtual try-on methods~~\cite{viton:18, gfla:2020, viton2:20} use deep neural networks to synthesize clothed humans in different poses. Sometimes, 2D segmentations~\cite{vtoseg:21} and body keypoints~\cite{pgpig:17} are used as auxiliary information to improve synthesizing quality. 3D-based virtual try-on~\cite{3dvto:22,smplicit:21,dg:17, dgr:16} reconstruct a
3D garment from image and re-target it to 3D human. The main challenges lie in solving human-cloth collision~\cite{3dvto:22} and representing a 3D garment with high fidelity to its texture and topology~\cite{3dvto2:20}.

\noindent\textbf{Human Mesh Recovery (HMR).}
Human mesh recovery aims at recovering 3D human mesh from person image. Many works employ parametric mesh model~\cite{smpl:15,smplify:19,star:20} driven by pose and shape parameters as 3D human representation. Regression based methods use CNN~\cite{hmr:18,expose:20}, transformer~\cite{meshtransformer:21,prtr:21} or GNN~\cite{meshgraphormer:21,pose2mesh:20,cmr:19} to extract features from images and estimate mesh parameters. Optimization-based methods~\cite{smplify:19, eft:20, spin:19} recover body pose by optimizing the projected mesh joints towards 2D ground-truth keypoints. HybrIK~\cite{hybrik:21} combines deep learning methods with analytical inverse kinematics methods to estimate pose parameters. However, in cases where large occlusions are present, multiple feasible meshes may correspond to a single image. Therefore, some methods~\cite{prohmr:21, multibodies:20} model the mesh parameters as a probabilistic distribution conditioned on the input image to address this challenge.

\noindent\textbf{Shape Under Clothing.} 
Estimating shape under clothing is challenging due to loose clothing types, body occlusions and depth ambiguity. SIZER~\cite{sizer:20}, ClothCap~\cite{clothcap:17} register 3D models to clothed scans to construct shape data. ~\cite{naketruth:08, em:16, zhang:17} uses motion sequence or multiple views to improve shape estimation. ~\cite{Shapy:22} also uses measurements and semantic segmentation to guide shape estimation, but cases where the human body is invisible are still an open problem.

%
%

\section{Our Approach}

\begin{figure*}[h]
\centering
\includegraphics[width=0.95\textwidth]{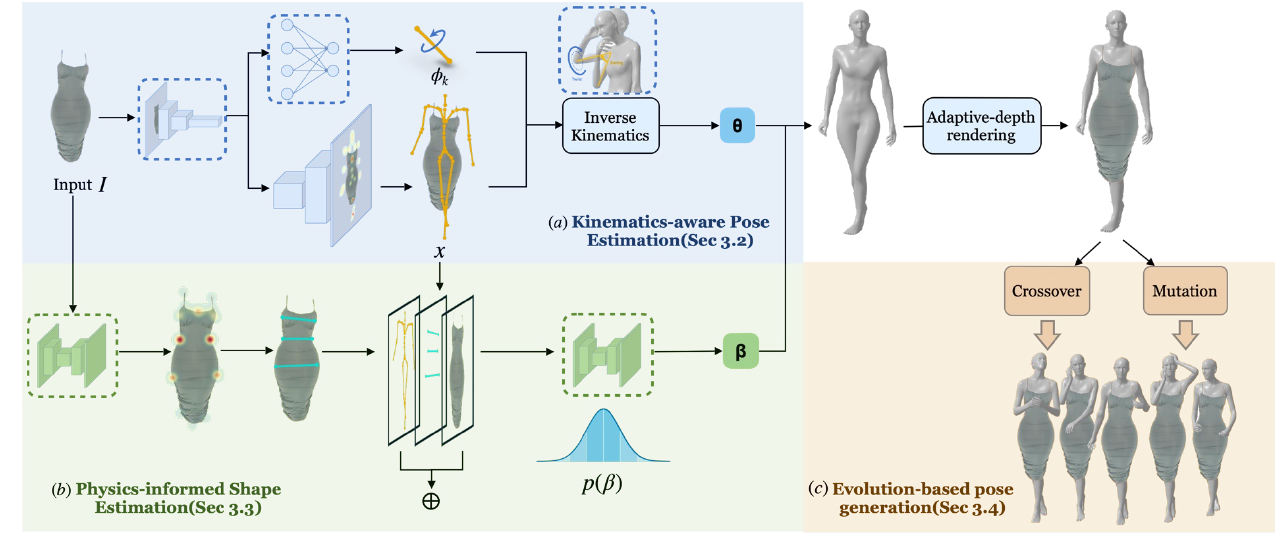} 
\vspace{-10pt}
\caption{\textbf{Framework Overview}. Our framework consists of three components. {\bf(a)} The kinematics-aware pose estimation module explicitly estimates 3D joint $x$ and bone twist angle $\phi$ from $I$ using a neural network, followed by an inverse kinematics module to calculate $\theta$. {\bf(b)} The physics-informed shape estimation branch estimates explicit body measurements from image and regress the $\beta$ distribution based on them. {\bf(c)} The evolution-based pose generation module produces diverse cloth-conditioned poses by KNN matching and skeleton transplanting.
}
\label{fig2}
\end{figure*}

The proposed method (see Figure~\ref{fig2}) consists of 3 components: kinematics-aware pose estimation branch, physics-informed shape estimation branch and an evolution-based pose generation branch. In the following sections, we will first introduce some preliminary methods upon which our work is built. Then, we will introduce the three components in our framework.


\noindent\textbf{Preliminary.}
Given a clothing image $I \in R^{H \times W \times C}$, our goal is to generate a 3D body mesh $M$ whose 2D projection fits into the clothing well. We choose SMPL~\cite{smpl:15}, a parametric body model to represent the generated 3D body mesh. SMPL model is a triangulated mesh $M \in R^{V\times3}$ with $V=6890$ vertices controlled by two sets of variables, pose parameters  $\theta \in R^n$ and shape parameters $\beta \in R^m$.  $\theta$ is an $n$-dimensional variable used to control body poses, where $n$ is the number of body joints and ${\theta}_i$ is the relative rotation matrix of the $i_{th}$ joint regarding to its parent in the skeleton tree. $\beta$ is used to describe body shape variation with predefined number of $m$ principle components in parameter space.  In this setup, our goal can be decomposed into three parts: estimating cloth part pose $\theta_{c}$, estimating $\beta$, and modeling invisible body pose $P(\theta_{o})$.

\subsection{Kinematics-aware Pose Estimation}
Our pose branch takes the clothing image $I$ as input and produces pose parameters $\theta$. Following HybrIK~\cite{hybrik:21}, we first estimate pixel-aligned postion of joints $x_k \in R^{3}$ and bone twist angle ${\phi}_k \in R$ for each joint $k\in\{0...n\}$. Then, we conduct inverse kinematics (IK) to analytically calculate the rotation matrix ${\theta}_k$ for each joint. This kinematics-aware approach binds recovered mesh to observable clothing area joints, resulting in significantly better reconstruction accuracy and robustness compared to parameter-regression based methods, as shown in Figure~\ref{fig:comparison}.
\subsubsection{3D Joints Estimation}
In our Cloth2Body setting, body mesh $M$ can be divided into two parts, $M_c$ and $M_o$. $M_c$ denotes the mesh area covered by clothes and $M_o$ is the body area outside clothes, \ie, $M=M_c \cup M_o$. We observe that $M_c$ is almost deterministic and can be estimated from $I$ while $M_o$ should be modeled as stochastic. We first use an encoder with ResNet34~\cite{resnet:16} backbone to extract features from $I$, followed by a deconvolution head producing 3D joints heatmaps $H$, and an MLP head producing bone twist angle ${\phi}$ of $J$. Finally, pixel-aligned joint coordinates $\hat x$ are calculated as the expectation of $H$. 
We use weighted L1 loss to supervise $x$ and a weighted L2 loss for $\phi$, where we assign lower weight $w_k$ to invisible $x_k \in M_o$ and higher $w_k$ to those visible $x_k \in M_c$:
\begin{equation}
\mathcal{L}_{kp} = \sum_{k=0}^{n} w_k \| x_k -\hat{x}_k \|_1
\end{equation}
\begin{equation}
\mathcal{L}_{tw} = \sum_{k=0}^{n} w_k \|(\cos{\phi_k}, \sin{\phi_k}) - (\cos\hat{\phi}_k, \sin\hat{\phi}_k) \|_2
\end{equation}

With 3D coordinates $x$ and bone twist angles $\phi$, we will apply inverse kinematics on joints to get $\theta$, described next.


\subsubsection{Inverse Kinematics}

Inverse kinematics (IK) takes $x$ and $\phi$ as input and analytically calculates joint rotation matrix $\theta$. This scheme is first introduced to pose estimation by HybrIK ~\cite{hybrik:21} and achieved near SOTA accuracy without much tuning. We retarget this method to our setting because it can adhere the recovered body to our estimated pixel-aligned joints through inverse kinematics. The key insight is decomposing ${\theta}$ into twist \ie, axial direction rotation matrix $\theta_{tw}$ and swing, \ie, radial direction rotation matrix $\theta_{sw}$ as: 
\begin{equation}
\theta = \theta_{sw}\theta_{tw},
\end{equation} 
where  $\theta_{tw}$ is converted from $\phi$ estimated by neural network on last step and $\theta_{sw}$ is analytically computed from $x$ by IK. As illustrated in Figure~\ref{fig2}(a), we denote the resting bone vector as $\vec t$ and target bone vector as $\vec p$. Then,  $\theta_{tw}$ is the rotation along $\vec{t}$ and $\theta_{sw}$ is the rotation around axis $\vec{n}$, which can be calculated as:
\begin{equation}
\vec n = \frac{\vec t \times \vec p}{\| \vec t \times \vec p\|}.
\end{equation}
Then, the swing angle $\alpha$ can be calculated as:
\begin{equation}
    \cos\alpha = \frac{\vec{t} \cdot \vec{p}}{\| \vec t \|\| \vec p\|},
    \sin\alpha = \frac{\| \vec t \times \vec p\|}{\| \vec t \|\| \vec p\|}.
\end{equation}
Then, we can obtain $\theta_{sw}$ using the \textit{Rodrigues Formula}:
\begin{equation}
\theta_{sw} = I + (\sin\alpha) K + (1-\cos\alpha)K^2,
\end{equation}where K is the cross-product matrix for the unit vector $\vec n$. 

\subsubsection{Adaptive Depth Estimation}

One key reason that the re-projected body cannot align well with the image is the camera estimation error. Human shape and camera depth are entangled in projection. For example, a taller man standing farther would appear the same in 2D image as a shorter man standing nearer to camera lens. Besides, IK would accumulate error if two sets of skeletons have different bone lengths, as shown in~\cite{hybrik:21}. However, in HybrIK, camera depth and shape parameters are predicted separately at inference, breaking their dependency with each other. 
To this end, we propose adaptive depth estimation trick. We first recover $M$ with shape $\beta$ and template pose. We then select a subset of anchor skeleton bones and calculate their lengths $\{b_{cam}^i | 1 <  i  \leq k\}$ at camera space from $M$ and $\{ b_{img}^i | 1 <  i  \leq k\}$ at image space from $x$. The camera depth $z_{cam}$ can be computed by skeleton size ratio:
\begin{equation}
\frac{z_{cam}}{f} = \frac{\sum_i^n{b_{cam}^i}}{\sum_i^n{b_{img}^i}},
\end{equation}where $f$ is the focal length of a perspective camera.


\subsection{Physics-informed Shape Estimation}
\begin{figure}
    \centering
    \includegraphics[width=0.8\linewidth]{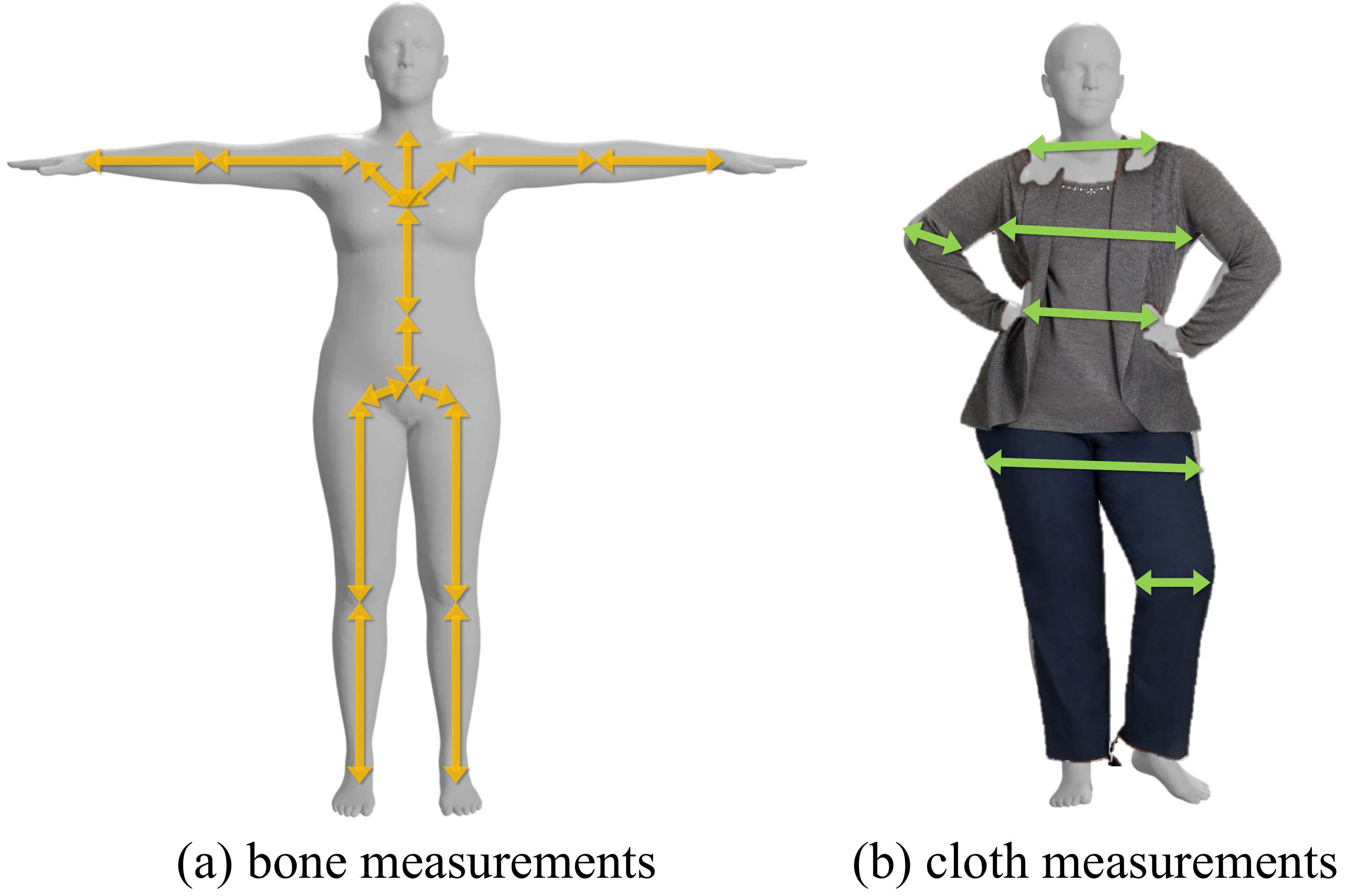}
    \caption{\fontsize{8.5pt}{8.5pt}\selectfont Illustration of body measurements. (a) shows the axial body measurements, \ie, bone lengths. (b) shows the radial body measurements such as shoulder width and breast width, which are estimated from clothing landmarks or input by users.} 
    \label{fig:measurements}
\end{figure}
In our shape estimation branch, we estimate clothing landmarks from the input clothing image $I_{rgb}$. Along with body joints estimated from pose branch, we calculate both axial and radial body measurements and render them as measurements skeleton image $I_{rgb}$, $I_{\hat{jts}}$ and $I_{\hat{lmk}}$ as illustrated in Figure~\ref{fig:measurements}. We then estimate SMPL shape parameters $\beta$ conditioned on measurements and pelvis depth using a probabilistic model. Our idea of explicitly estimating cloth landmarks and body joints as a physical proxy is based on the observation that while clothing image potentially embeds body shape information, it's challenging to accurately estimate body shape from a single clothing image due to occlusions, variations in clothing types and depth ambiguities~\cite{shape_under_cloth:08, shape_under_cloth:17}. Results in Table~\ref{tb:shape} show that our physics-informed shape estimation can achieve significantly better shape estimation results.

\label{subsec:m2s}

\subsubsection{Keypoints to Measurements}

Our keypoints set includes body joints and cloth landmarks. As shown in Figure~\ref{fig:measurements}, we first define a set of physical measurements $\omega$ consisting of axial direction measurements $\omega_{ax}$ and radial direction measurements $\omega_{rd}$, \ie, $\omega = \omega_{ax} \cup \omega_{rd}$. $\omega_{ax}$ is responsible for attributes such as height, leg length and arm span, calculated using skeleton $x$. $\omega_{rd}$ influences attributes regarding body weight such as breast and hip width, calculated from detected clothing landmarks. To be specific, we define a mapping to align landmarks of different cloth categories in DeepFashion2~\cite{df2:19} based on their semantic meaning so that the uniform body measurements set $\omega$ can be calculated consistently from landmarks and joints. For example, the shoulder width measurement can be estimated using the distance between the left and right shoulder landmarks if available for input clothing types. 

Note that despite we use estimated landmarks and joints to calculate $\omega$, they can also be provided by clothing manufacturers or interactively manipulated by users. Some manipulation results are shown in Figure~\ref{fig:manipulate_shape}.

\subsubsection{Probabilistic Shape Estimation}

Stochasticity is an inherent challenge in our Cloth2Body setting. Shape under clothing can have multiple feasible solutions. For example, the shape of body parts not covered by clothing is undecided; the shape of body in loose clothing is not strictly constrained; and 3D height is highly entangled with camera distance from only 2D observation. The noises of landmark detection also exacerbate this problem. To this end, we model $\beta$ as a sample from conditional distribution $p_\theta(\beta | z)$,  where $z$ is the latent feature extracted from $I_{rgb}$, $I_{\hat{jts}}$ and $I_{\hat{lmk}}$ using ResNet~\cite{resnet:16} concatenated with camera distance $d_{cam}$, which in inference time is estimated by RootNet~\cite{rootnet:19}:
\begin{equation}
\begin{aligned}
& I_{\hat{jts}} = \pi(f_1(I_{rgb})), \space I_{\hat{lmk}} = \pi(f_2(I_{rgb})), \\
& z = f(I_{rgb} \oplus I_{\hat{jts}} \oplus I_{\hat{lmk}}) \oplus d_{cam},
\end{aligned}
\end{equation}
where $d_{cam}$ is the estimated distance from camera to human body, $\pi$ is the render operation, $f_1$ is the pose estimation model and $f_2$ is the landmark detection model.
The extracted feature is then mapped to a latent distribution $q_\phi$, which is learned using ELBO~\cite{vae:13}: 

\begin{equation}
\mathcal{L}_{m2s} = -\mathbb{E}_{z \sim q_\phi}\left[\log \frac{p_{\theta}(\beta,z)}{q_{\phi}(z)}\right]
\end{equation}

\subsection{Evolution-based Pose Generation}
\begin{figure}
    \centering
    \includegraphics[width=1.0\linewidth]{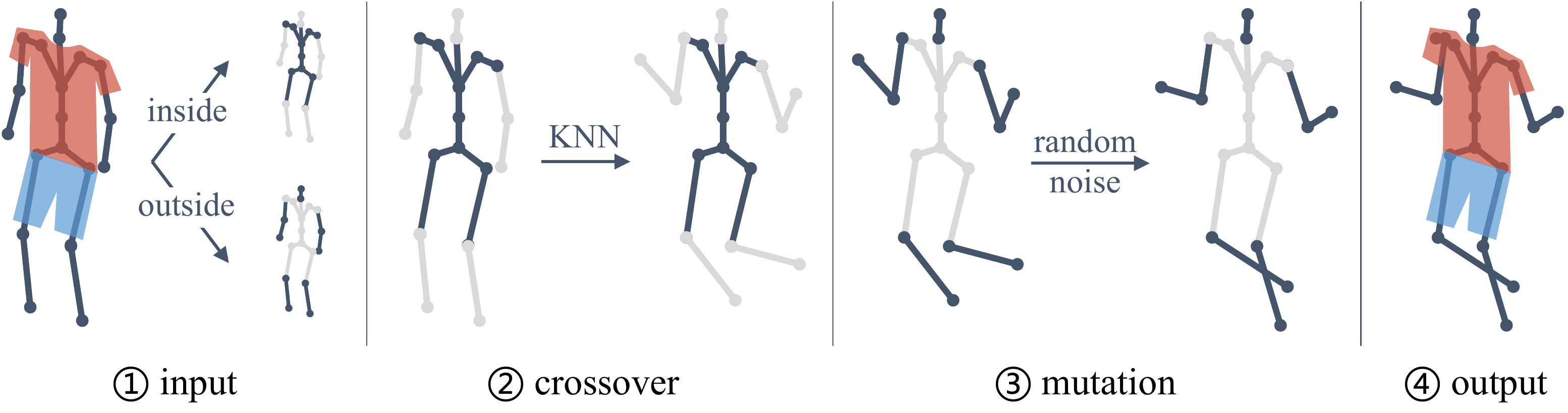}
    \caption{\fontsize{8.5pt}{8.5pt}\selectfont In crossover, we transplant compatible skeletons in database to the invisible part of estimated skeleton. In mutation, we disturb the rotation angle in a small range to diversify poses.}
    \label{fig:crossover-mutation}
\end{figure}
Our pose generation module aims to diversify body poses conditioned on clothes. Inspired by evolutionary algorithms~\cite{evoskeleton:20}, we use crossover and mutation to evolve skeletons and poses in designated area. In pose crossover, we collect $\theta$ from real world dataset to form database $D_p$. The rotation of joint $i$ is annotated as ${\theta}_k^c$ if $i \in M_c$ elsewise ${\theta}_k^o$. During inference, we match $\theta_c$ with poses in $D_p$ by KNN to find reasonable surrogate $\theta^o$. The crossover operation in pose estimation can be formulated as:
\begin{equation}
\theta_{cross} = [\theta^c, \theta^o_k], \quad  k = \underset{k \in D_p}{\mathrm{argmin}}\|\theta^c - \theta^c_k\|.
\end{equation}
As for pose mutation, we randomly disturb the estimated $\theta^o$ with small noise to get diversified poses. The mutation operation is formulated as:
\begin{equation}
\theta_{mut} = [\theta^c, \theta^o + \Delta\theta^o], \quad \|\Delta\theta^o \|< \epsilon.
\end{equation}

With pose crossover and mutation, we can easily model the diversity of out-of-cloth body gestures with robustness, as shown in the right part of Figure~\ref{fig:comparison}.

\begin{table*}[ht]
\small
\centering

\resizebox{\linewidth}{!}{
\begin{tabular}{c|ccc|ccc|ccc}

\toprule
& \multicolumn{3}{c|}{AGORA} & \multicolumn{3}{c|}{DeepFashion2} & \multicolumn{3}{c}{Multi-Garment Net} \\
\cline{2-4} \cline{5-7} \cline{8-10}
& MPJPE-C $\downarrow$ & PA-MPJPE-C $\downarrow$ & 2D-KPE-C $\downarrow$ & MPJPE-C $\downarrow$ & PA-MPJPE-C $\downarrow$ & 2D-KPE-C $\downarrow$ & MPJPE-C $\downarrow$ & PA-MPJPE-C $\downarrow$ & 2D-KPE-C $\downarrow$\\
\midrule
HMR~\cite{hmr:18} & 83.51 & 63.44 & 31.68 & 87.88 & 59.63 & 37.27 & 109.73 & 80.79 & 24.04\\
PARE~\cite{pare:21} & 84.38 & 69.63 & 26.52 & 72.80 & 55.90 & 28.69 & 88.96 & 74.59 & 13.97\\
HybrIK~\cite{hybrik:21} & 60.49 & 51.83 & 19.47 & 65.16 & 43.82 & 27.46 & 88.71 & 74.22 & 11.87\\
Ours & \bf60.01 & \bf 51.48 & \bf 19.28 & \bf 62.35 & \bf 43.54 & \bf 23.53 & \bf 86.99 & \bf 73.93 & \bf 10.66 \\

\bottomrule
\end{tabular}
}
\caption{Clothing area joints accuracy on AGORA-CLOTH, DeepFashion2-SMPL and MGN dataset. We compare our methods to HMR, PARE and HybrIK and achieve the best cloth-human alignment results.}
\label{tb:main}
\end{table*}


\section{Experiments}
We evaluate our model performance qualitatively and quantitatively on synthetic dataset and real world dataset\footnote{More results and implementation details are reported in the supplementary materials.}.
\begin{table}[t]
\centering
\caption{Shape estimation accuracy. HMR, PARE and HybrIK directly regress $\beta$ from the image, while our method first estimates landmarks and joints as middle proxy.}
\vspace{-10pt}
\resizebox{\columnwidth}{!}{
\begin{tabular}{c|c|c|c|c}
\toprule
& Height (mm) $\downarrow$ & Chest (mm)$\downarrow$& Waist (mm)$\downarrow$ & Hips (mm)$\downarrow$\\
\midrule
HMR~\cite{hmr:18} & 76.0 & 20.0 & 22.6 & 21.0 \\
PARE~\cite{hybrik:21} & 79.3 & 64.8 & 63.2 & 28.7 \\
HybrIK~\cite{hybrik:21} & 78.3 & 20.3 & 23.2 & 21.0 \\
Ours & 75.8 & 18.7 & 21.3 & 19.6 \\
\bottomrule
\end{tabular}
}
\label{tb:shape}
\end{table}
\subsection{Datasets and Settings}
\label{subseg:dataset}
To the best of our knowledge, there exists no dataset containing both clothing images and their corresponding clothed human with 3D parametric annotations. To evaluate methods on the proposed Cloth2Body task, we introduce two new datasets, AGORA-CLOTH and Deepfashion2-SMPL, based on two existing datasets. \\
{\bf AGORA-CLOTH.}
We introduce a new synthetic dataset AGORA-CLOTH, which includes various cloth types and 3D body annotations. This dataset is adapted from AGORA~\cite{agora:21}, a synthetic 3D human body dataset with accurate 3D annotations and rich clothing types. However, AGORA does not provide clothing images required as inputs in our setting. Thus, We crop the clothing part from original images using segmentation methods. Images with too much occlusion are dropped during processing. The final collected AGORA-CLOTH training dataset includes $\sim$75,000 single-person cloth images and annotations. \\
{\bf DeepFashion2-SMPL.}
We also propose a real-world dataset DeepFashion2-SMPL which consists of tremendous fashionable clothing images and corresponding 3D human annotations. DeepFashion2~\cite{df2:19} (DF2) is a large image dataset containing real-world humans with fashionable poses and covers 13 types of clothes with annotated clothing landmarks. To adapt DF2 dataset to Cloth2Body, we fit SMPL~\cite{smpl:15} on DeepFashion2 dataset combining regression-based method~\cite{hybrik:21} and optimization-based method~\cite{smplify:19}. After fitting, we filter the results by fitting errors and keep the well-reconstructed ones in our final dataset. The final collected dataset includes $\sim$8,700 single-person images with 3D annotations. This Deepfashion2-SMPL dataset is used to evaluate our Cloth2Body framework. \\
{\bf Multi-Garment Net.} Multi-Garment net(MGN)~\cite{mgn:19} provides 3D clothed humans as well as their SMPL registrations. We get the clothing image by first rendering the groundtruth meshes at an average distance and then applying a segmentation algorithm to crop the clothing part out. We only use MGN for evaluation.

\subsection{Evaluation Metrics}
\noindent{\bf Pose Evaluation.}
To evaluate pose estimation quality in cloth area, we propose two new metrics: mean per joint position error in cloth(MPJPE-C) and procrustes aligned MPJPE in cloth(PA-MPJPE-C), adapted from MPJPE and PA-MPJPE, two commonly used metrics in 3D pose estimation.
MPJPE-C computes the mean position error of joints lied in clothing area. Since only joints in this area are almost deterministic from the given clothing image, higher MPJPE-C can indicate better cloth-conditioned pose estimation quality. PA-MPJPE-C is a variant of MPJPE-C, which eliminates the effect of translation, rotation and scaling via procrustes alignment.\\
\noindent{\bf Shape Evaluation.}
To evaluate body shape estimation quality, we compute typical anthropometric measurements of the output 3d mesh and calculate their difference from ground-truth value. Specifically, we measure height and the width of chest, waist and hips similar to SHAPY~\cite{Shapy:22}. 
We choose width instead of circumference used in SHAPY so that we can seamlessly use 2D distance between clothing landmarks to estimate body measurements at inference.\\
\noindent{\bf Image Alignment Evaluation.}
The pixel-wise alignment between the re-projected human mesh and clothing image is very important for downstream applications such as reshaping and pose transfer. To evaluate pixel-wise alignment quality, we use 2D keypoint error(2D-KPE) which computes the average distance between re-projected body joints and ground-truth body joints in 2D image space.

\begin{figure*}[ht]
  \centering
  \includegraphics[width=0.95\linewidth]{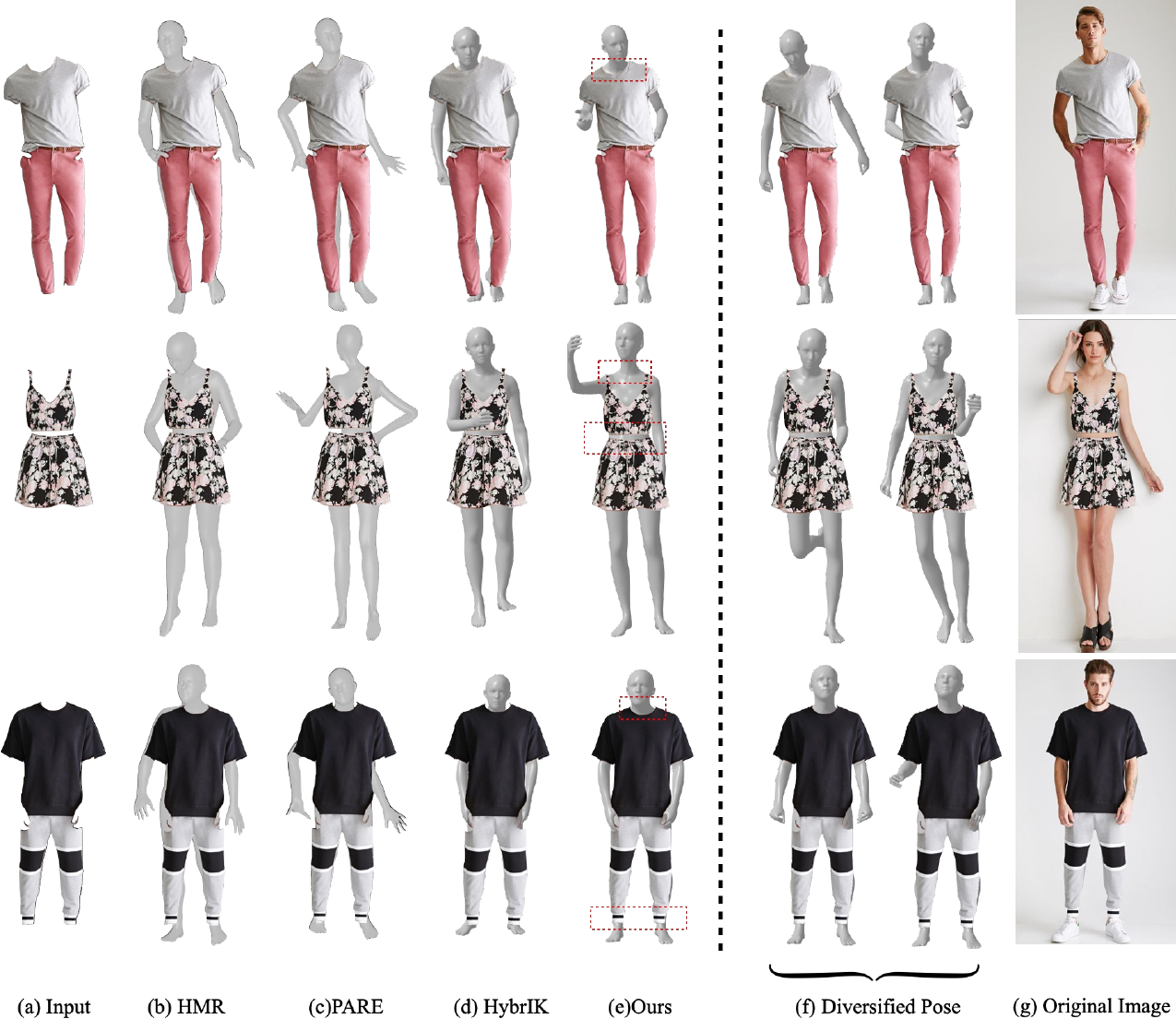}
  \caption{\fontsize{8.5pt}{8.5pt}\selectfont Visualization results of generated meshes of several methods and our diverse pose generation method.}
  \label{fig:comparison}
\end{figure*}

\subsection{Comparative Studies}

Since this is a new task setting with no existing end-to-end solution, we adapted three widely-used methods in human mesh recovery tasks to our task for comparison, \ie, HMR~\cite{hmr:18}, PARE~\cite{pare:21} and HybrIK~\cite{hybrik:21}. Although many other works ~\cite{smplify:19, spin:19, expose:20} have been introduced to human mesh recovery, they cannot be trivially adapted to our task since they need other information such as human keypoints. Our implementation is based on mmhuman3d~\cite{mmhuman3d} framework. All these baselines are trained from scratch on AGORA-CLOTH dataset.\\
{\bf Body Alignment.}
We compared our method with baselines on AGORA-CLOTH and DeepFashion2-SMPL dataset about their reconstruction quality and cloth-human alignment accuracy. 
Our method achieves the most accurate body joints alignment (see Table~\ref{tb:main}) and the most natural clothing detail alignment (see Figure~\ref{fig:comparison}). HMR and PARE can generate reasonable poses conditioned on clothes and roughly align recovered mesh to the clothing area, but they lack naturality in generated poses and robustness to clothing types. 
HybrIK can generate a better-aligned human mesh and more natural pose in our setting, but still cannot align well with clothes in detail, especially clothing edges. More importantly, all these methods  directly regress $\beta$ from RGB information and  tend to generate average-shaped human, which further hurts cloth-human alignment.\label{subsug:shape_exp}\\
{\bf Shape Estimation.}
We evaluate the performance of our Landmark2Shape scheme on DeepFashion2-SMPL dataset. As shown in Table~\ref{tb:shape}, our model achieves the best accuracy in shape estimation in all testing indicators, including height, chest, waist and hips, which means the recovered $\beta$ in our pipeline successfully captures the human figure underneath a clothing image. \\
\noindent{\bf Diverse Body Generation.}
In the right half of Figure~\ref{fig:comparison}, we display several diversified poses that fit into the same clothing. During post-processing, we transplant poses only to body parts not covered by clothing. Results show that our model can generate reasonable and diverse poses conditioned on clothing. For example, the arms and hands can freely cross, bend or naturally hang down along body side, while keeping a good alignment between the whole body and the clothing. To test the effectiveness of our defined measurements to control $\beta$, we train an MLP regressor that maps measurements to shape and visualize the results of measurement editing in Figure~\ref{fig:manipulate_shape}.

\begin{figure*}[ht]
    \centering
    \includegraphics[width=\linewidth]{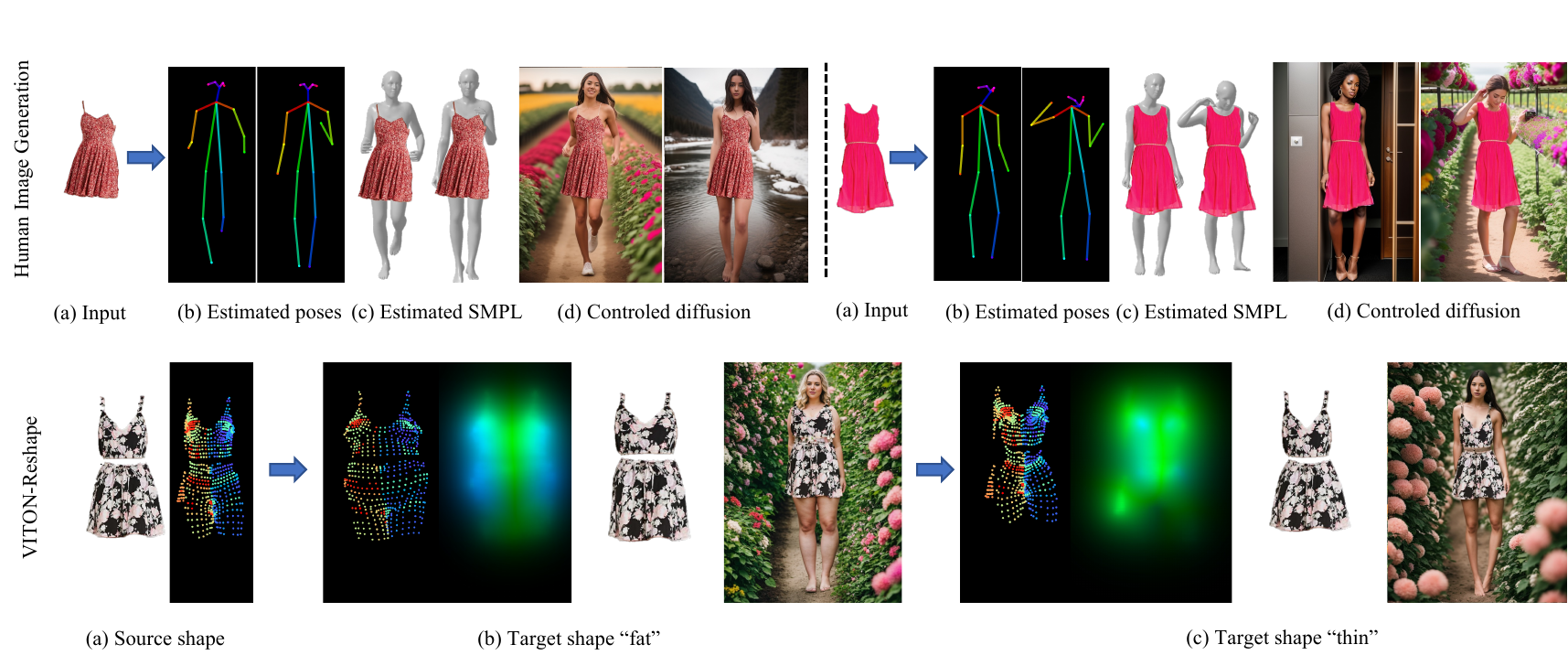}
    \vspace{-20pt}
    \caption{\fontsize{8.5pt}{8.5pt}\selectfont Application cases of our Cloth2Body setting. The upper line demonstrates the human image generation task with (a)two input samples. (b)(c)Our model generates 3D skeleton and mesh given a single clothing image, which is consumed by (d) a stable diffusion model to generate a photorealistic human image. The lower line demonstrates a clothing reshape task in virtual try-on. (a) is the clothing image input and the 3D SMPL vertices extracted from our Cloth2Body estimation. We then can manipulate $\beta$ to obtain (b)reshaped meshes and dense warp field generated from 3D correspondence, and generate models of different shapes. (c) is the same reshaping pipeline but set $\beta$ as "thin".}
    \label{fig:sc}
\end{figure*}

\subsection{Applications}

\begin{figure}
    \centering
    \includegraphics[width=\linewidth]{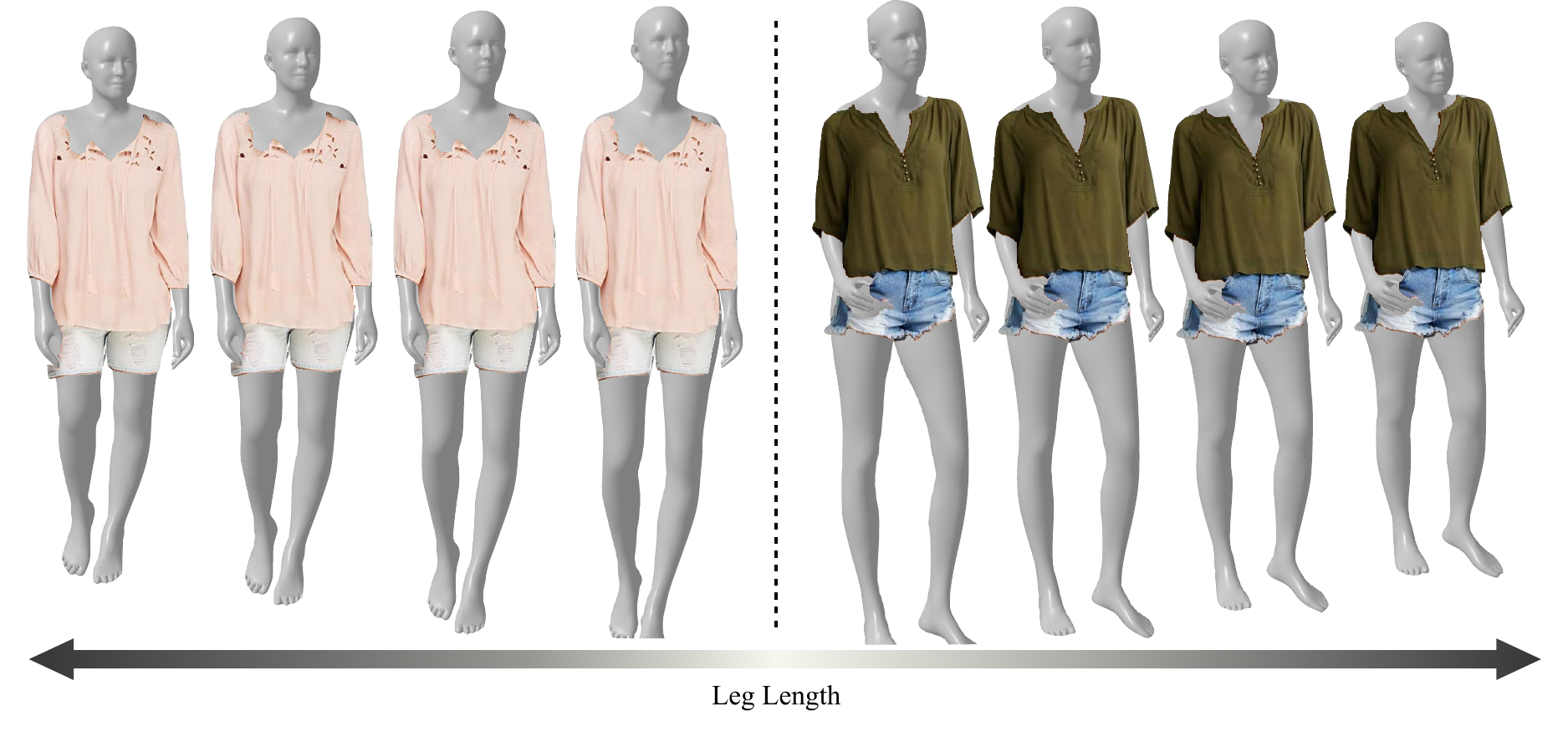}
    \vspace{-10pt}
    \caption{\fontsize{8.5pt}{8.5pt}\selectfont Measurements editing. We modify the leg length entry in $\omega$ and show that mesh shape varies accordingly.}
    \label{fig:manipulate_shape}
    \vspace{-10pt}
\end{figure}
\
We show two application scenarios here. The first is human image generation. Traditional human image generation methods often require a reference human as input and would fail to generate a real-like human with only clothing image. Instead, our model generates 3D meshes and pose skeletons that are well-aligned to the clothing in the image space. This serves as a strong guiding signal for generative models to generate both controllable and diverse human images. In Figure~\ref{fig:sc}, we extend our pipeline with a pretrained stable diffusion module~\cite{controlnet:23}, which generates photo-realistic human images guided by our 3D outputs.

The second application is virtual try-on(VITON). VITON tasks~\cite{viton:18} involve re-targeting clothing images to human models of various poses and shapes. They often compute 2D warp fields based on 2D correspondences extracted from image, which often lacks 3D priors when transforming images. Since our model can generate pixel-aligned and parametric mesh from clothing images, we can estimate the clothing warp field using the dense 3D correspondences of human mesh vertices. We demonstrate a clothing reshape application in Figure~\ref{fig:sc}. By manipulating SMPL shape parameters, we easily achieve natural and fine-grained control of clothing shape and size.


\subsection{Ablation Studies}

\noindent{\bf Adaptive Depth.}
To examine how adaptive depth aids 3D pose estimation, we only enable the adaptive depth module and evaluate 2D-KPE on DeepFashion2-SMPL dataset. Inference time depth is estimated via RootNet~\cite{rootnet:19} following HybrIK. As shown in Table~\ref{tb:ablation_adaptive_depth}, both HybrIK and our model with adaptive depth achieve lower 2D alignment error. The adaptive depth module results in better 2D alignment not only because it makes camera depth aware of body shape, but also because it aids the inverse kinematic process by aligning joints better in space.

\begin{table}[h]
\centering
\caption{Ablation on adaptive depth. Results show it improves 2D body-cloth alignment in both HybrIK and ours.}
\vspace{-10pt}
\footnotesize
\begin{tabular}{c|c|c|c}
\toprule
 & 2D-KPE-C $\downarrow$ & & 2D-KPE-C $\downarrow$\\
\midrule
HybrIK~\cite{hybrik:21} & 31.58 & Ours & 23.53 \\
HybrIK-Ada  & \bf27.46 & Ours-Ada & \bf 25.72\\

\bottomrule
\end{tabular}
\label{tb:ablation_adaptive_depth}
\end{table}



\noindent{\bf Landmark2Shape.}
To investigate the contribution of each input signal to the shape estimation module, we trained two controlled-group models separately. The first model was trained using only the RGB image $I_{rgb}$, while the second was trained with both the $I_{rgb}$ and the self-predicted joints image, $I_{\hat{jts}}$. Our full model utilized all three input signals, including $I_{rgb}$, $I_{\hat{jts}}$, and $I_{\hat{lmk}}$. As shown in Table~\ref{tb:ablation_l2s}, $I_{\hat{jts}}$ and $I_{\hat{lmk}}$ both contribute to shape estimation.

\begin{table}[h]
\centering
\caption{Ablation on landmarks2shape. We test shape accuracy w/o explicitly estimated landmarks and joints.}
\vspace{-10pt}
\resizebox{0.9\columnwidth}{!}{
\begin{tabular}{c|c|c|c|c|c}
\toprule
& $I_{rgb}$ & $I_{\hat{jts}}$ & $I_{\hat{lmk}}$ & MPJPE-C $\downarrow$ & 2D-KPE-C $\downarrow$ \\
\midrule
control\_1 & $\checkmark$ & - & - & 63.67 & 23.76 \\
control\_2 & $\checkmark$ & $\checkmark$ & - & 62.76 & 23.56 \\
full & $\checkmark$ & $\checkmark$ & $\checkmark$ & 62.35 & 23.53 \\
\bottomrule
\end{tabular}
}
\label{tb:ablation_l2s}
\end{table}


\section{Discussion and Conclusion}
\noindent \textbf{Discussion.}
There is still room to improve this work. In challenging non-frontal viewpoints, the predicted 3D mesh may fail to align with the 2D clothes accurately. Besides, measurements computed from cloth landmarks are not robust across rare clothing types. To address these limitations, we need better semantic understanding of clothes.\\
\noindent \textbf{Conclusion.}
In this paper, we introduced a novel Cloth2Body task which aims to generate multiple plausible 3D human body meshes from a single 2D clothing image. Specifically, we designed a two-branch framework which estimates the pose and shape parameters of a parametric 3D body model. Also, we designed an adaptive depth trick and a Landmark2Shape method for pixel-wise alignment and fine-grained body shape control. Moreover, our method allows users to generate diverse natural poses via our evolution-based pose generation method.  Finally, qualitative and quantitative evaluations on real-world data demonstrate the superiority of the proposed framework over other alternatives.\\
\noindent \textbf{Broad Impact.}
We found that human dataset lacks variety in shape and may bias towards thinner humans. For better performance and fairness, more diversified 3D human models are expected.

\section{Acknowledgement}
This research was supported in part by the National Natural Science Foundation of China under Grant No.62102110, Foshan HKUST Projects (FSUST21-FYTRI01A, FSUST21-FYTRI02A), the Ministry of Education, Singapore, under its MOE AcRF Tier 2 (MOE-T2EP20221-0012), NTU NAP, and under the RIE2020 Industry Alignment Fund – Industry Collaboration Projects (IAF-ICP) Funding Initiative, as well as cash and in-kind contribution from the industry partner(s).

{\small
\bibliographystyle{ieee_fullname}
\bibliography{egbib}

\begin{thebibliography}{10}\itemsep=-1pt

\bibitem{naketruth:08}
A. Balan and M.~J. Black.
\newblock The naked truth: Estimating body shape under clothing,.
\newblock In {\em European Conf. on Computer Vision, ECCV}, volume 5304, pages 15--29, 2008.

\bibitem{shape_under_cloth:08}
A. Balan and M.~J. Black.
\newblock The naked truth: Estimating body shape under clothing,.
\newblock In {\em European Conference on Computer Vision (ECCV)}, volume 5304, pages 15--29, 2008.

\bibitem{mgn:19}
Bharat~Lal Bhatnagar, Garvita Tiwari, Christian Theobalt, and Gerard Pons-Moll.
\newblock Multi-garment net: Learning to dress 3d people from images.
\newblock In {\em {IEEE} International Conference on Computer Vision ({ICCV})}. {IEEE}, Oct 2019.

\bibitem{multibodies:20}
Benjamin Biggs, S{\'{e}}bastien Ehrhart, Hanbyul Joo, Benjamin Graham, Andrea Vedaldi, and David Novotny.
\newblock {3D} multibodies: Fitting sets of plausible {3D} models to ambiguous image data.
\newblock In {\em Conference on Neural Information Processing Systems (NeurIPS)}, 2020.

\bibitem{eg3d:22}
Eric~R. Chan, Connor~Z. Lin, Matthew~A. Chan, Koki Nagano, Boxiao Pan, Shalini~De Mello, Orazio Gallo, Leonidas Guibas, Jonathan Tremblay, Sameh Khamis, Tero Karras, and Gordon Wetzstein.
\newblock Efficient geometry-aware {3D} generative adversarial networks.
\newblock In {\em {IEEE} Conference on Computer Vision and Pattern Recognition (CVPR)}, pages 16102--16112, 2022.

\bibitem{pose2mesh:20}
Hongsuk Choi, Gyeongsik Moon, and Kyoung~Mu Lee.
\newblock Pose2mesh: Graph convolutional network for 3d human pose and mesh recovery from a 2d human pose.
\newblock In {\em European Conference on Computer Vision (ECCV)}, pages 769--787, 2020.

\bibitem{Shapy:22}
Vasileios Choutas, Lea M{\"{u}}ller, Chun-Hao~P. Huang, Siyu Tang, Dimitris Tzionas, and Michael~J. Black.
\newblock Accurate 3d body shape regression using metric and semantic attribute.
\newblock In {\em {IEEE} Conference on Computer Vision and Pattern Recognition (CVPR)}, pages 2708--2718, 2022.

\bibitem{expose:20}
Vasileios Choutas, Georgios Pavlakos, Timo Bolkart, Dimitrios Tzionas, and Michael~J. Black.
\newblock Monocular expressive body regression through body-driven attention.
\newblock In {\em European Conference on Computer Vision (ECCV)}, 2020.

\bibitem{mmhuman3d}
MMHuman3D Contributors.
\newblock Openmmlab 3d human parametric model toolbox and benchmark.
\newblock \url{https://github.com/open-mmlab/mmhuman3d}, 2021.

\bibitem{smplicit:21}
Enric Corona, Albert Pumarola, Guillem Aleny{\`a}, Gerard Pons-Moll, and Francesc Moreno-Noguer.
\newblock Smplicit: Topology-aware generative model for clothed people.
\newblock In {\em IEEE Conference on Computer Vision and Pattern Recognition (CVPR)}, pages 11875--11885, 2021.

\bibitem{dg:17}
R. Dan\'{z}\v{r}ek, E. Dibra, C. \"{O}ztireli, R. Ziegler, and M. Gross.
\newblock Deepgarment: 3d garment shape estimation from a single image.
\newblock {\em Computer Graphics Forum (CGF)}, page 269–280, 2017.

\bibitem{vr:06}
Philippe Decaudin, Dan Julius, Jamie Wither, Laurence Boissieux, Alla Sheffer, and Marie-Paule Cani.
\newblock Virtual garments: A fully geometric approach for clothing design.
\newblock In {\em Computer Graphics Forum (CGF)}, pages 625--634, 2006.

\bibitem{vtoseg:21}
Yuying Ge, Yibing Song, Ruimao Zhang, Chongjian Ge, Wei Liu, and Ping Luo.
\newblock Parser-free virtual try-on via distilling appearance flows.
\newblock {\em arXiv preprint arXiv:2103.04559}, 2021.

\bibitem{df2:19}
Yuying Ge, Ruimao Zhang, Lingyun Wu, Xiaogang Wang, Xiaoou Tang, and Ping Luo.
\newblock A versatile benchmark for detection, pose estimation, segmentation and re-identification of clothing images.
\newblock {\em {IEEE} Conference on Computer Vision and Pattern Recognition (CVPR)}, pages 5337--5345, 2019.

\bibitem{viton:18}
Xintong Han, Zuxuan Wu, Zhe Wu, Ruichi Yu, and Larry~S Davis.
\newblock Viton: An image-based virtual try-on network.
\newblock In {\em IEEE Conference on Computer Vision and Pattern Recognition (CVPR)}, pages 7543--7552, 2018.

\bibitem{mae:21}
Kaiming He, Xinlei Chen, Saining Xie, Yanghao Li, Piotr Doll{\'a}r, and Ross Girshick.
\newblock Masked autoencoders are scalable vision learners.
\newblock {\em arXiv preprint arXiv:2111.06377}, 2021.

\bibitem{resnet:16}
Kaiming He, Xiangyu Zhang, Shaoqing Ren, and Jian Sun.
\newblock Deep residual learning for image recognition.
\newblock In {\em Proceedings of the IEEE Conference on Computer Vision and Pattern Recognition (CVPR)}, pages 770--778, 2016.

\bibitem{garment4d:21}
Fangzhou Hong, Liang Pan, Zhongang Cai, and Ziwei Liu.
\newblock Garment4d: Garment reconstruction from point cloud sequences.
\newblock In {\em Conference on Neural Information Processing Systems (NeurIPS)}, pages 27940--27951, 2021.

\bibitem{avatarclip:22}
Fangzhou Hong, Mingyuan Zhang, Liang Pan, Zhongang Cai, Lei Yang, and Ziwei Liu.
\newblock Avatarclip: Zero-shot text-driven generation and animation of 3d avatars.
\newblock {\em ACM Transactions on Graphics}, 41(4):1--19, 2022.

\bibitem{eft:20}
Hanbyul Joo, Natalia Neverova, and Andrea Vedaldi.
\newblock Exemplar fine-tuning for 3d human pose fitting towards in-the-wild 3d human pose estimation.
\newblock In {\em International Conference on 3D Vision (3DV)}, pages 42--52, 2020.

\bibitem{hmr:18}
Angjoo Kanazawa, Michael~J. Black, David~W. Jacobs, and Jitendra Malik.
\newblock End-to-end recovery of human shape and pose.
\newblock In {\em {IEEE} Conference on Computer Vision and Pattern Recognition (CVPR)}, pages 7122--7131, 2018.

\bibitem{vae:13}
Diederik~P Kingma and Max Welling.
\newblock Auto-encoding variational bayes.
\newblock {\em arXiv preprint arXiv:1312.6114}, 2013.

\bibitem{pare:21}
Muhammed Kocabas, Chun-Hao~P. Huang, Otmar Hilliges, and Michael~J. Black.
\newblock {PARE}: Part attention regressor for {3D} human body estimation.
\newblock In {\em International Conference on Computer Vision (ICCV)}, pages 11127--11137, 2021.

\bibitem{spin:19}
Nikos Kolotouros, Georgios Pavlakos, Michael~J Black, and Kostas Daniilidis.
\newblock Learning to reconstruct 3d human pose and shape via model-fitting in the loop.
\newblock In {\em International Conference on Computer Vision (ICCV)}, pages 2252--2261, 2019.

\bibitem{cmr:19}
Nikos Kolotouros, Georgios Pavlakos, and Kostas Daniilidis.
\newblock Convolutional mesh regression for single-image human shape reconstruction.
\newblock In {\em {IEEE} Conference on Computer Vision and Pattern Recognition (CVPR)}, pages 4501--4510, 2019.

\bibitem{prohmr:21}
Nikos Kolotouros, Georgios Pavlakos, Dinesh Jayaraman, and Kostas Daniilidis.
\newblock Probabilistic modeling for human mesh recovery.
\newblock In {\em International Conference on Computer Vision (ICCV)}, 2021.

\bibitem{hybrik:21}
Jiefeng Li, Chao Xu, Zhicun Chen, Siyuan Bian, Lixin Yang, and Cewu Lu.
\newblock Hybrik: A hybrid analytical-neural inverse kinematics solution for 3d human pose and shape estimation.
\newblock In {\em {IEEE} Conference on Computer Vision and Pattern Recognition (CVPR)}, pages 3383--3393, 2021.

\bibitem{prtr:21}
Ke Li, Shijie Wang, Xiang Zhang, Yifan Xu, Weijian Xu, and Zhuowen Tu.
\newblock Pose recognition with cascade transformers.
\newblock In {\em {IEEE} Conference on Computer Vision and Pattern Recognition (CVPR)}, pages 1944--1953, 2021.

\bibitem{evoskeleton:20}
Shichao Li, Lei Ke, Kevin Pratama, Yu-Wing Tai, Chi-Keung Tang, and Kwang-Ting Cheng.
\newblock Cascaded deep monocular 3d human pose estimation with evolutionary training data.
\newblock In {\em {IEEE} Conference on Computer Vision and Pattern Recognition (CVPR)}, 2020.

\bibitem{meshtransformer:21}
Kevin Lin, Lijuan Wang, and Zicheng Liu.
\newblock End-to-end human pose and mesh reconstruction with transformers.
\newblock In {\em {IEEE} Conference on Computer Vision and Pattern Recognition (CVPR)}, pages 1954--1963, 2021.

\bibitem{meshgraphormer:21}
Kevin Lin, Lijuan Wang, and Zicheng Liu.
\newblock Mesh graphormer.
\newblock In {\em International Conference on Computer Vision (ICCV)}, pages 12919--12928, 2021.

\bibitem{dflmk:16}
Ziwei Liu, Sijie Yan, Ping Luo, Xiaogang Wang, and Xiaoou Tang.
\newblock Fashion landmark detection in the wild.
\newblock In {\em European Conference on Computer Vision (ECCV)}, pages 229--245, 2016.

\bibitem{smpl:15}
Matthew Loper, Naureen Mahmood, Javier Romero, Gerard Pons-Moll, and Michael~J. Black.
\newblock {SMPL}: A skinned multi-person linear model.
\newblock {\em ACM Transactions on Graphics}, 34(6):248:1--248:16, Oct. 2015.

\bibitem{pgpig:17}
Liqian Ma, Xu Jia, Qianru Sun, Bernt Schiele, Tinne Tuytelaars, and Luc Van~Gool.
\newblock Pose guided person image generation.
\newblock In {\em Conference on Neural Information Processing Systems (NeurIPS)}, pages 405--415, 2017.

\bibitem{3dvto2:20}
Qianli Ma, Jinlong Yang, Anurag Ranjan, Sergi Pujades, Gerard Pons-Moll, Siyu Tang, and Michael~J Black.
\newblock Learning to dress 3d people in generative clothing.
\newblock In {\em {IEEE} Conference on Computer Vision and Pattern Recognition (CVPR)}, pages 6468--6477, 2020.

\bibitem{rootnet:19}
Gyeongsik Moon, Juyong Chang, and Kyoung~Mu Lee.
\newblock Camera distance-aware top-down approach for 3d multi-person pose estimation from a single rgb image.
\newblock In {\em International Conference on Computer Vision (ICCV)}, pages 10132--10141, 2019.

\bibitem{nie2021pose2room}
Yinyu Nie, Angela Dai, Xiaoguang Han, and Matthias Nie{\ss}ner.
\newblock Pose2room: Understanding 3d scenes from human activities.
\newblock In {\em {IEEE} Conference on Computer Vision and Pattern Recognition (CVPR)}, pages 425--443, 2021.

\bibitem{star:20}
Ahmed A.~A. Osman, Timo Bolkart, and Michael~J. Black.
\newblock {STAR}: A sparse trained articulated human body regressor.
\newblock In {\em European Conference on Computer Vision (ECCV)}, pages 598--613, 2020.

\bibitem{agora:21}
Priyanka Patel, Chun-Hao~P. Huang, Joachim Tesch, David~T. Hoffmann, Shashank Tripathi, and Michael~J. Black.
\newblock {AGORA}: Avatars in geography optimized for regression analysis.
\newblock In {\em {IEEE} Conference on Computer Vision and Pattern Recognition (CVPR)}, pages 13468--13478, 2021.

\bibitem{smplify:19}
Georgios Pavlakos, Vasileios Choutas, Nima Ghorbani, Timo Bolkart, Ahmed A.~A. Osman, Dimitrios Tzionas, and Michael~J. Black.
\newblock Expressive body capture: 3d hands, face, and body from a single image.
\newblock In {\em {IEEE} Conference on Computer Vision and Pattern Recognition (CVPR)}, pages 10975--10985, 2019.

\bibitem{clothcap:17}
Gerard Pons-Moll, Sergi Pujades, Sonny Hu, and Michael~J Black.
\newblock Clothcap: Seamless 4d clothing capture and retargeting.
\newblock {\em ACM Transactions on Graphics (ToG)}, 36(4):1--15, 2017.

\bibitem{dreamfusion:22}
Ben Poole, Ajay Jain, Jonathan~T. Barron, and Ben Mildenhall.
\newblock Dreamfusion: Text-to-3d using 2d diffusion.
\newblock {\em arXiv preprint arXiv:2209.14988}, 2022.

\bibitem{gfla:2020}
Yurui Ren, Ge Li, Shan Liu, and Thomas~H Li.
\newblock Deep spatial transformation for pose-guided person image generation and animation.
\newblock {\em IEEE Transactions on Image Processing}, 29:8622--8635, 2020.

\bibitem{vto:19}
Igor Santesteban, Miguel~A Otaduy, and Dan Casas.
\newblock Learning-based animation of clothing for virtual try-on.
\newblock In {\em Computer Graphics Forum}, pages 355--366, 2019.

\bibitem{3dvto:22}
Igor Santesteban, Miguel~A Otaduy, Nils Thuerey, and Dan Casas.
\newblock Ulnef: Untangled layered neural fields for mix-and-match virtual try-on.
\newblock In {\em Conference on Neural Information Processing Systems (NeurIPS)}, 2022.

\bibitem{garmentcollision:21}
Igor Santesteban, Nils Thuerey, Miguel~A Otaduy, and Dan Casas.
\newblock {Self-Supervised Collision Handling via Generative 3D Garment Models for Virtual Try-On}.
\newblock {\em {IEEE} Conference on Computer Vision and Pattern Recognition (CVPR)}, pages 11763--11773, 2021.

\bibitem{hmrreview:22}
Yating Tian, Hongwen Zhang, Yebin Liu, and Limin Wang.
\newblock Recovering 3d human mesh from monocular images: A survey.
\newblock {\em arXiv preprint arXiv:2203.01923}, 2022.

\bibitem{sizer:20}
Garvita Tiwari, Bharat~Lal Bhatnagar, Tony Tung, and Gerard Pons-Moll.
\newblock Sizer: A dataset and model for parsing 3d clothing and learning size sensitive 3d clothing.
\newblock In {\em European Conference on Computer Vision ({ECCV})}, August 2020.

\bibitem{wang2021scene}
Jingbo Wang, Sijie Yan, Bo Dai, and Dahua Lin.
\newblock Scene-aware generative network for human motion synthesis.
\newblock In {\em IEEE Conference on Computer Vision and Pattern Recognition (CVPR)}, pages 12206--12215, 2021.

\bibitem{viton2:20}
Han Yang, Ruimao Zhang, Xiaobao Guo, Wei Liu, Wangmeng Zuo, and Ping Luo.
\newblock Towards photo-realistic virtual try-on by adaptively generating-preserving image content.
\newblock In {\em {IEEE} Conference on Computer Vision and Pattern Recognition (CVPR)}, pages 7847--7856, 2020.

\bibitem{em:16}
Jinlong Yang, Jean-S{\'e}bastien Franco, Franck H{\'e}troy-Wheeler, and Stefanie Wuhrer.
\newblock Estimation of human body shape in motion with wide clothing.
\newblock In Bastian Leibe, Jiri Matas, Nicu Sebe, and Max Welling, editors, {\em European Conf. on Computer Vision, ECCV}, pages 439--454, 2016.

\bibitem{dgr:16}
Shan Yang, Tanya Amert, Zherong Pan, Ke Wang, Licheng Yu, Tamara~L. Berg, and Ming~C. Lin.
\newblock Detailed garment recovery from a single-view image.
\newblock {\em arXiv preprint arXiv:1608.01250}, 2016.

\bibitem{yi2022human}
Hongwei Yi, Chun-Hao~P Huang, Dimitrios Tzionas, Muhammed Kocabas, Mohamed Hassan, Siyu Tang, Justus Thies, and Michael~J Black.
\newblock Human-aware object placement for visual environment reconstruction.
\newblock In {\em IEEE Conference on Computer Vision and Pattern Recognition (CVPR)}, pages 3949--3960, 2022.

\bibitem{zhang:17}
Chao Zhang, Sergi Pujades, Michael Black, and Gerard Pons-Moll.
\newblock Detailed, accurate, human shape estimation from clothed {3D} scan sequences.
\newblock In {\em 2017 IEEE Conference on Computer Vision and Pattern Recognition (CVPR)}, pages 5484--5493, 2017.
\newblock Spotlight.

\bibitem{shape_under_cloth:17}
Chao Zhang, Sergi Pujades, Michael Black, and Gerard Pons-Moll.
\newblock Detailed, accurate, human shape estimation from clothed {3D} scan sequences.
\newblock In {\em 2017 IEEE Conference on Computer Vision and Pattern Recognition (CVPR)}, pages 5484--5493, 2017.

\bibitem{controlnet:23}
Lvmin Zhang and Maneesh Agrawala.
\newblock Adding conditional control to text-to-image diffusion models, 2023.

\bibitem{comodgan:21}
Shengyu Zhao, Jonathan Cui, Yilun Sheng, Yue Dong, Xiao Liang, Eric~I Chang, and Yan Xu.
\newblock Large scale image completion via co-modulated generative adversarial networks.
\newblock In {\em International Conference on Learning Representations (ICLR)}, 2021.

\end{thebibliography}
}

\end{document}